\documentclass[lettersize,journal]{IEEEtran}

\usepackage{amsmath,amsfonts}
\usepackage{algorithm}
\usepackage{algpseudocode}
\usepackage{array}
\usepackage[caption=false,font=normalsize,labelfont=sf,textfont=sf]{subfig}
\usepackage{textcomp}
\usepackage{url}
\usepackage{verbatim}
\usepackage{graphicx}
\usepackage{cite}
\usepackage{booktabs}
\usepackage{dsfont}
\usepackage{tabularx}
\usepackage{dblfloatfix} 
\usepackage{makecell}
\usepackage{multirow}
\usepackage{float} 
\usepackage{cuted}
\usepackage{capt-of}
\usepackage{orcidlink}

\begin{document}

\title{Granular Ball Guided Masking:\\ Structure-aware Data Augmentation}

\author{%
    Shuyin~Xia\orcidlink{0000-0001-5993-9563},~
    Fan~Chen\orcidlink{0009-0007-2163-5557},~
    Dawei~Dai*\orcidlink{0000-0002-8431-4431},~
    Meng~Yang\orcidlink{0009-0005-8311-092X},~
    Junwei~Han,~\IEEEmembership{Fellow,~IEEE},~
    
    Xinbo~Gao\orcidlink{0000-0002-7985-0037},~\IEEEmembership{Fellow,~IEEE},~
    and Guoyin~Wang\orcidlink{0000-0002-8521-5232},~\IEEEmembership{Senior~Member,~IEEE}%
    \thanks{This work was supported by Chongqing Key Laboratory of Computational Intelligence, 
    Key Laboratory of Cyberspace Big Data Intelligent Security, Ministry of Education, 
    Sichuan-Chongqing Co-construction Key Laboratory of Digital Economy Intelligence, 
    and Key Laboratory of Big Data Intelligent Computing, 
    Chongqing University of Posts and Telecommunications.}%
    \thanks{*Corresponding author: Dawei Dai (dw\_dai@163.com), 
    Chongqing University of Posts and Telecommunications, China.}%
}


\markboth{Journal of \LaTeX\ Class Files,~Vol.~14, No.~8, October~2025}%
{Shell \MakeLowercase{\textit{et al.}}: A Sample Article Using IEEEtran.cls for IEEE Journals}


\maketitle

\begin{abstract}
Deep learning models have achieved remarkable success in computer vision but still rely heavily on large-scale labeled data and tend to overfit when data is limited or distributions shift.  
Data augmentation---particularly mask-based information dropping---can enhance robustness by forcing models to explore complementary cues; however, existing approaches often lack structural awareness and risk discarding essential semantics.  
We propose Granular Ball Guided Masking (GBGM), a structure-aware augmentation strategy guided by Granular Ball Computing (GBC).  
GBGM adaptively preserves semantically rich, structurally important regions while suppressing redundant areas through a coarse-to-fine hierarchical masking process, producing augmentations that are both representative and discriminative.  
Extensive experiments on multiple benchmarks demonstrate consistent improvements not only in image classification and masked image reconstruction, but also in image tampering detection, validating the effectiveness and generalization of GBGM across both recognition and forensic scenarios.  
Simple and model-agnostic, GBGM integrates seamlessly into CNNs and Vision Transformers, offering a practical paradigm for structure-aware data augmentation.
\end{abstract}


\section{Introduction}

Deep learning models—ranging from Convolutional Neural Networks (CNNs) to the more advanced Vision Transformers (ViTs)—have achieved breakthrough progress in image classification~\cite{krizhevsky2012imagenet,he2016deep,dosovitskiy2020image}, object detection~\cite{ren2015faster,redmon2016you,he2017mask}, and semantic segmentation~\cite{long2015fully,ronneberger2015u,chen2017deeplab}.  
Despite these successes, such models contain millions of parameters and depend heavily on large-scale labeled data, making them prone to overfitting~\cite{huang2017densely} and poor generalization when data is limited or the distribution shifts~\cite{hendrycks2019robustness,geirhos2019imagenet}.  
Data augmentation serves as a fundamental strategy to expand training distributions and improve robustness.

\begin{figure}[t]
  \centering
  \includegraphics[width=0.9\linewidth]{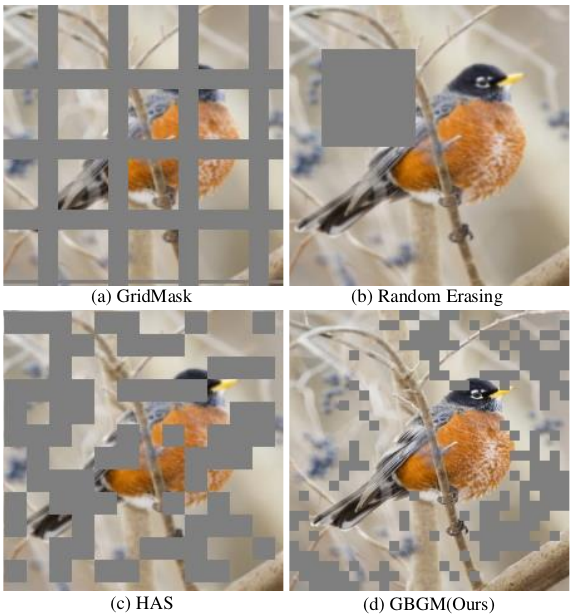}
\caption{Comparison of four augmentation strategies: 
(a) GridMask, (b) Random Erasing, (c) saliency-based HAS, and (d) our GBGM.}
  \label{fig:example}
\end{figure}

Data augmentation methods can be grouped into spatial transformations~\cite{krizhevsky2012imagenet}, color distortions~\cite{szegedy2015going}, and information dropping~\cite{devries2017improved}.  
The first two simulate natural appearance changes, while information-dropping approaches—such as Random Erasing~\cite{zhong2020random}, GridMask~\cite{chen2020gridmask}, and Hide-and-Seek (HaS)~\cite{singh2018hide}—intentionally remove or transform image regions to encourage complementary feature learning and reduce over-reliance on local cues.  
However, these methods face a persistent trade-off: too much deletion discards critical semantics, whereas too little yields trivial gains, and their random or rigid patterns lack structural awareness, making it difficult to balance preservation with effective perturbation.

Granular Ball Computing (GBC) has recently emerged as a lightweight paradigm for structural, multi-granularity modeling, demonstrating strong capability in adaptively partitioning visual space~\cite{wang2017dgcc,Xia2023GranularballCA,cheng2024gb,xie2024gbg++,quadir2024granular,10996538}.  
However, current data augmentation approaches largely lack explicit structural modeling and therefore struggle to preserve global-to-local semantic consistency.  
By contrast, GBC naturally provides structural awareness and multi-scale representation, offering the desirable ability to identify and retain informative regions while filtering redundant background.  
This observation motivates us to leverage GBC as the foundation for a principled and hierarchical augmentation mechanism.

As illustrated in Fig.~\ref{fig:example}, GBGM provides a clear contrast to existing augmentation strategies such as GridMask, Random Erasing, and HAS. While these baselines remove image content in random or heuristic ways, GBGM adaptively retains informative content while filtering redundant regions, producing masks that integrate naturally into CNNs or ViTs.

Building on these insights, our contributions are threefold:
\begin{itemize}
  \item \textbf{Structure-aware masking.} Unifies Granular Ball Computing with saliency-driven erasure to produce masks that are both discriminative and structure-preserving.
  \item \textbf{Hierarchical mask generation.} Employs a multi-stage coarse-to-fine procedure that balances global representativeness with fine-grained discriminability for robust augmentations.
  \item \textbf{Lightweight, model-agnostic design.} Requires no extra supervision or external models and can be seamlessly applied to diverse architectures and resource-constrained scenarios.
\end{itemize}

\section{Related Work}

Several research directions underpin our study.  
Data augmentation, including masking-based approaches, improves model robustness by diversifying training data, while Granular Ball Computing provides a lightweight and structurally aware representation that inspires our method.

\textbf{Data augmentation.}
Operating solely on input data, data augmentation enhances generalization without altering network structure or introducing auxiliary losses.  
Basic strategies such as random flipping and cropping are standard in CNNs.  
Building on these, Inception-preprocessing~\cite{szegedy2015going} adds random color perturbations, AutoAugment~\cite{cubuk2019autoaugment} searches for optimal augmentation policies.  
Information-dropping methods (e.g., Random Erasing~\cite{zhong2020random}, Hide-and-Seek (HaS)~\cite{singh2018hide}, GridMask~\cite{chen2020gridmask}) further remove regions to encourage complementary cues.  
These techniques establish a rich foundation for improving generalization across scales and model families.  
Yet their masking strategies are largely random or heuristic, motivating the need for more structure-aware and adaptive approaches.

\textbf{Masking-based augmentations.}
Within data augmentation, a prominent family explicitly removes or masks local image regions.  
Among them, Random Erasing~\cite{zhong2020random} stochastically removes rectangular areas, AutoAugment~\cite{cubuk2019autoaugment} discovers effective masking and transformation policies, GridMask~\cite{chen2020gridmask} imposes regular grid occlusions, and Hide-and-Seek (HaS)~\cite{singh2018hide} randomly hides patches at multiple scales to diversify learned features.  
These approaches are computationally lightweight but rely on random or heuristic patterns, which may over-delete critical semantics or provide limited regularization.  
They typically yield strong gains on small-scale benchmarks like CIFAR-10/100 but bring only modest improvements on large-scale datasets where structural cues dominate.  
In contrast, our GBGM introduces a structure-aware masking mechanism guided by Granular Ball representations, bridging the gap between block-level occlusion and information-driven regularization for more stable and stronger gains.

\textbf{Granular Ball Computing.}
Human cognition follows the principle of “global precedence”~\cite{chen1982topological}.  
Based on traditional granular computing, Wang et al.~\cite{wang2017dgcc} proposed multi-granular cognitive computing, and Xia et al.~\cite{Xia2023GranularballCA} further introduced Granular Ball Computing (GBC), which represents and covers sample space with balls of varying sizes for multi-scale learning.  
Granular Ball Computing has been applied across various artificial intelligence domains,  examples include Granular Ball reinforcement learning \cite{liu2024unlock}, Granular Ball classifiers \cite{xia2022efficient, xia2025graph}, Granular Ball clustering \cite{xia2020ball,xie2024mgnr, xie2024gbg++}, Granular Ball support vector machines \cite{quadir2024granular}, and Granular Ball graph representation\cite{10996538}.
Building on this concept, we develop GBGM, which leverages multi-scale Granular Ball representations to guide hierarchical mask generation with strong structural awareness and low computational cost.

\section{Method}

Given an input image of size $H \times W$, GBGM progressively filters regions by local purity, propagating only the most informative areas and yielding a final binary mask $\mathcal{M}_{\text{final}}$ of size $H \times W$. At its core, Granular balls—compact structural units rather than individual pixels—capture local properties and provide the purity metric guiding selection.  

\begin{figure*}[t]
  \centering
  \includegraphics[width=0.9\textwidth]{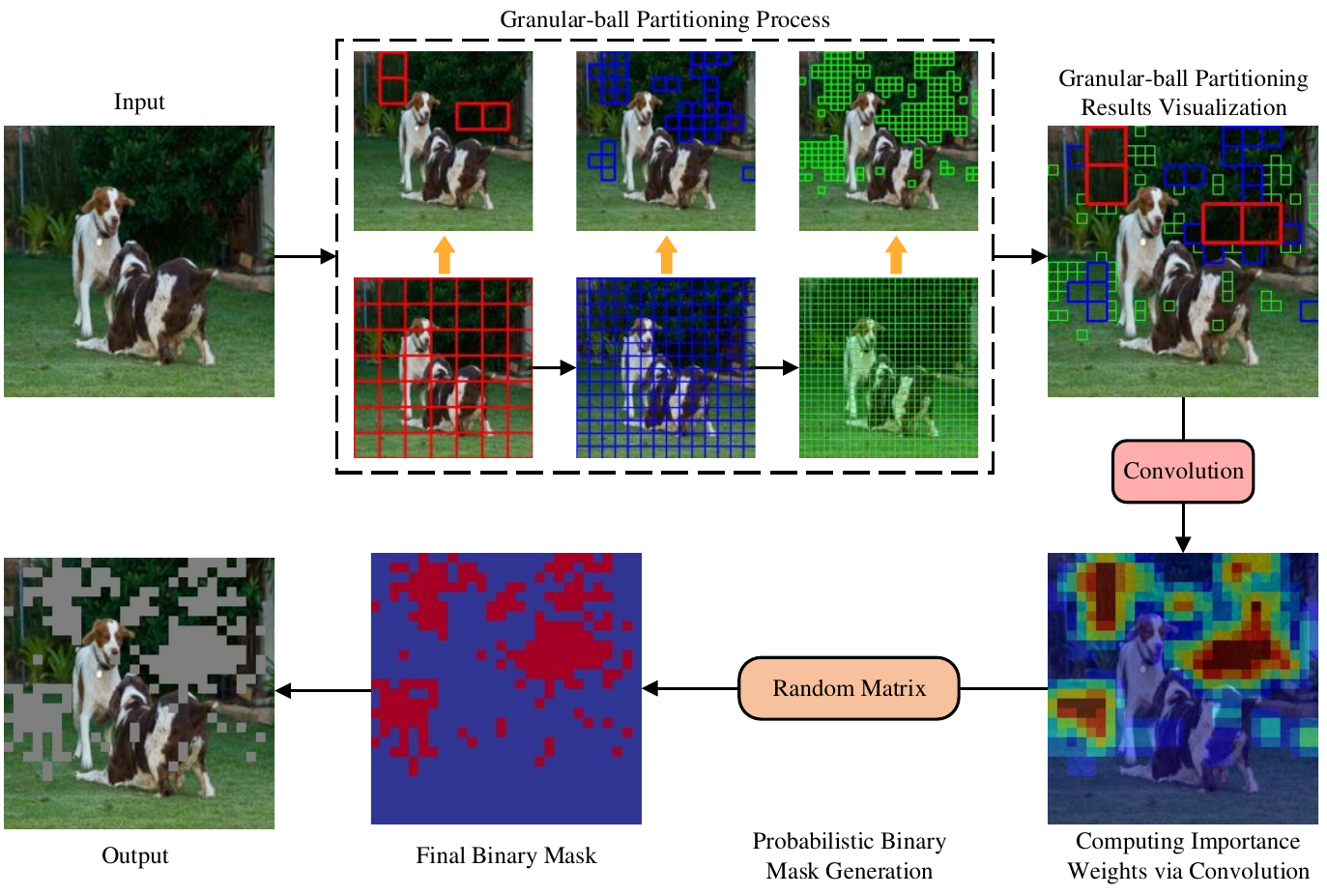}
  \caption{Overview of the proposed GBGM pipeline. Granular Ball analysis guides a hierarchical selection process to retain informative regions, producing a high-resolution binary importance mask.}
  \label{fig:system}
\end{figure*}

Formally, given a dataset $\mathcal{D}$, a Granular Ball is denoted as $\mathcal{GB}_j$ $(j=1,2,\ldots,k)$, where $k$ is the total number of Granular balls and $\mathcal{GB}_j=\{\, y_i \mid 1 \leq i \leq m_j \,\}$ contains $m_j$ data points. The center of $\mathcal{GB}_j$ is defined as $c_j=\tfrac{1}{m_j}\sum_{i=1}^{m_j}y_i$, and its radius as $r_j=\tfrac{1}{m_j}\sum_{i=1}^{m_j}\lvert y_i-c_j\rvert$. When the quality of a Granular Ball fails to meet the threshold $T$, it is recursively split into smaller Granular balls until each satisfies $T$. This iterative refinement ensures convergence and yields clearer decision boundaries. In GBGM, these Granular balls serve as the structural primitives for hierarchical mask generation, as illustrated in Fig.~\ref{fig:system}.

\subsection{Stage 1: Coarse Granularity Masking}

The input image of size $H \times W$ is first divided into a grid of non-overlapping square blocks, each with side length $S_1$.  
In practice, $S_1$ is chosen proportional to the input resolution to maintain a similar semantic granularity.  
For example, we use $S_1=4$ for $32 \times 32$ inputs (single-stage masking) and $S_1=32$ for $224 \times 224$ inputs, where a second-stage refinement with $S_2=16$ is applied.  
Other resolutions can follow the same scaling principle by choosing $S_1$ proportional to the input size, if needed.  
The number of blocks along height and width are given by:
\begin{equation}
N_{1}^{(h)} = \frac{H}{S_1}, \quad N_{1}^{(w)} = \frac{W}{S_1},
\label{eq:coarse-grid-size}
\end{equation}
yielding a total number of coarse blocks:
\begin{equation}
N_1 = N_{1}^{(h)} \times N_{1}^{(w)}.
\label{eq:coarse-total}
\end{equation}

We define a coarse binary mask $\mathcal{M}_1 \in \{0,1\}^{N_{1}^{(h)} \times N_{1}^{(w)}}$ to represent retained blocks. Each block $\mathcal{P}_{i,j}$ is evaluated using a local purity score, quantifying the deviation of its central patch from the block's mean intensity:
\begin{equation}
\operatorname{Purity}(\mathcal{P}_{i,j}) = 
\frac{1}{|\mathcal{C}|} \sum_{(u,v) \in \mathcal{C}} 
\bigl| \mathcal{P}_{i,j}(u,v) - \mu(\mathcal{P}_{i,j}) \bigr|,
\label{eq:purity-score}
\end{equation}
where $\mathcal{C}$ is a small central patch (e.g., $2 \times 2$ pixels), and $\mu(\mathcal{P}_{i,j})$ is the mean pixel intensity of block $\mathcal{P}_{i,j}$.

We then select the top-$k_1$ blocks with the highest purity scores. Formally, letting $\tau_1$ denote the purity threshold corresponding to the $k_1$-th largest purity score, the coarse mask is defined by
\begin{equation}
\mathcal{M}_1(i,j) = 
\begin{cases}
1 & \text{if } \operatorname{Purity}(\mathcal{P}_{i,j}) \geq \tau_1, \\
0 & \text{otherwise}.
\end{cases}
\label{eq:topk-selection}
\end{equation}

This stage removes large, uninformative background regions while preserving semantically salient areas, providing an efficient approximation of high-content zones to guide subsequent refinement.

\subsection{Stage 2: Finer Granularity Masking}

To refine the selection, all blocks rejected by $\mathcal{M}_1$ (i.e., where $\mathcal{M}_1(i,j)=0$) are subdivided into $2 \times 2$ finer sub-blocks. The finer block size is
\begin{equation}
S_2 = \frac{S_1}{2},
\label{eq:fine-block-size}
\end{equation}
leading to finer grid resolutions
\begin{equation}
N_{2}^{(h)} = \frac{H}{S_2}, \quad N_{2}^{(w)} = \frac{W}{S_2}.
\label{eq:fine-grid-size}
\end{equation}

The total number of fine blocks is
\begin{equation}
N_2 = N_{2}^{(h)} \times N_{2}^{(w)} = 4N_1.
\label{eq:fine-total}
\end{equation}

For each fine block $\mathcal{Q}_{i',j'}$ originating from rejected coarse blocks, we compute the purity score as in Eq.~\eqref{eq:purity-score}. Denote by $\tau_2$ the threshold corresponding to the top-$k_2$ purest fine blocks; the finer binary mask $\mathcal{M}_2 \in \{0,1\}^{N_2^{(h)} \times N_2^{(w)}}$ is defined as
\begin{equation}
\mathcal{M}_2(i',j') = 
\begin{cases}
1 & \text{if } \operatorname{Purity}(\mathcal{Q}_{i',j'}) \geq \tau_2, \\
0 & \text{otherwise}.
\end{cases}
\label{eq:topk-fine}
\end{equation}

This stage recovers fine-grained structures like edges and small objects missed in the coarse stage, improving mask precision while limiting computation to previously discarded regions.

\subsection{Stage 3: Importance Mask Generation}

After obtaining the finer binary mask from Stage 2, we refine it by considering the spatial context of each block. Let the finer mask be $\mathcal{M}_2$ of size $N_2^{(h)} \times N_2^{(w)}$. Instead of simple smoothing, we perform a $3 \times 3$ convolution with an all-one kernel to count the number of neighboring blocks marked as target:
\begin{equation}
I = \operatorname{Conv}_{3 \times 3}(\mathcal{M}_2),
\label{eq:stage3_conv}
\end{equation}
where each element of $I$ represents the local importance of the corresponding block. Blocks located on object boundaries or heterogeneous regions receive higher scores, while homogeneous background areas have lower scores.

The importance map $I$ is normalized to $[0,1]$ to ensure numerical stability:
\begin{equation}
\hat{I} = \frac{I - \operatorname{min}(I)}{\operatorname{max}(I) - \operatorname{min}(I) + \varepsilon}.
\label{eq:stage3_normalization}
\end{equation}
to introduce stochasticity and promote spatial diversity, a random matrix $R$ is drawn from the uniform distribution $\mathcal{U}(\varepsilon, 1-\varepsilon)$, and the low-resolution binary mask is computed as
\begin{equation}
\mathcal{M}_\text{final}^{(\text{lowres})}(i) = 
\begin{cases}
1 & \text{if } \hat{I}(i) < R(i), \\
0 & \text{otherwise}.
\end{cases}
\label{eq:stage3_sampling}
\end{equation}

This stochastic thresholding prevents deterministic patterns while respecting the relative importance encoded in $\hat{I}$. In this final mask, $0$ corresponds to high-purity background blocks that are masked out, and $1$ corresponds to low-purity, target-relevant blocks that are retained. This ensures that the mask emphasizes informative regions while suppressing homogeneous background areas.

Finally, the low-resolution mask is upsampled to the original image size $H \times W$ using bilinear or nearest-neighbor interpolation:
\begin{equation}
\mathcal{M}_{\text{final}} = 
\operatorname{Interp}_{H \times W}\!\left( 
\mathcal{M}_{\text{final}}^{(\text{lowres})} 
\right).
\label{eq:stage3_upsample}
\end{equation}
yielding the full-resolution binary importance mask for downstream tasks such as masked image modeling or attention guidance.

\subsection{Time Complexity Analysis}

We analyze the computational complexity of GBGM across its three stages. In Stage~1, the input image of size $H \times W$ is partitioned into $N_1=\tfrac{H}{S_1}\cdot\tfrac{W}{S_1}$ coarse blocks. Since purity is computed on a constant-sized central patch, the cost is $O(N_1)=O(\tfrac{HW}{S_1^2})$, and top-$k_1$ selection adds at most a logarithmic factor. In Stage~2, all rejected coarse blocks are subdivided into $2 \times 2$ finer blocks, producing up to $N_2=4N_1=O(\tfrac{HW}{S_2^2})$. Purity computation and selection again take $O(N_2)$ to $O(N_2\log N_2)$. In Stage~3, the importance mask is refined via a $3 \times 3$ convolution, normalization, random thresholding, and final interpolation, which together cost $O(N_2+HW)$. Combining the three stages yields
\begin{equation}
O\!\left(
\frac{HW}{S_1^2}\log\frac{HW}{S_1^2} \;+\;
\frac{HW}{S_2^2}\log\frac{HW}{S_2^2} \;+\;
HW
\right).
\end{equation}
Since $S_1$ and $S_2$ scale proportionally with $H$ and $W$ (e.g., $S_1=H/7$, $S_2=H/14$), both $N_1$ and $N_2$ are constant with respect to image size, leaving the interpolation term $O(HW)$ as the dominant factor. Thus, GBGM operates in linear time with respect to the number of pixels, ensuring scalability to high-resolution inputs.

\begin{table}[h]
\centering
\caption{Complexity analysis of GBGM stages.}
\label{tab:time-complexity-stages}
\begin{tabular}{lcc}
\toprule
Stage & Main operation & Complexity \\
\midrule
Stage~1 & Purity on $N_1$ coarse blocks & $O(\tfrac{HW}{S_1^2})$ \\
Stage~2 & Purity on $N_2$ finer blocks & $O(\tfrac{HW}{S_2^2})$ \\
Stage~3 & Conv + normalization + upsample & $O(N_2+HW)$ \\
\midrule
Overall & Full GBGM pipeline & $O(HW)$ (linear) \\
\bottomrule
\end{tabular}
\end{table}

\begin{algorithm}[H]
\caption{Granular Ball Guided Masking Pipeline}
\label{alg:granular_ball_masking}
\begin{algorithmic}[1]
\Require Input image $X$ of size $H \times W$, coarse block size $S_1$, top-$k_1$, top-$k_2$, small constant $\varepsilon$
\Ensure Full-resolution binary mask $\mathcal{M}_{\text{final}} \in \{0,1\}^{H \times W}$

\State Compute coarse grid resolution: $N_1^{(h)} \gets H / S_1$, $N_1^{(w)} \gets W / S_1$
\State Partition $X$ into coarse blocks $\{\mathcal{P}_{i,j}\}$ of size $S_1 \times S_1$
\For{each coarse block $\mathcal{P}_{i,j}$}
    \State Compute $\operatorname{Purity}(\mathcal{P}_{i,j})$ using Eq.~\eqref{eq:purity-score}
\EndFor
\State Select top-$k_1$ blocks with highest purity to form coarse mask $\mathcal{M}_1$
\State Define finer block size: $S_2 \gets S_1 / 2$
\State Compute finer grid: $N_2^{(h)} \gets H / S_2$, $N_2^{(w)} \gets W / S_2$
\State Initialize $\mathcal{M}_2$ to zeros
\For{each $\mathcal{P}_{i,j}$ with $\mathcal{M}_1(i,j)=0$}
    \State Partition $\mathcal{P}_{i,j}$ into $2 \times 2$ finer blocks $\{\mathcal{Q}_{i',j'}\}$
    \For{each $\mathcal{Q}_{i',j'}$}
        \State Compute $\operatorname{Purity}(\mathcal{Q}_{i',j'})$
    \EndFor
\EndFor
\State Select top-$k_2$ finer blocks to form $\mathcal{M}_2$
\State Compute $I \gets \operatorname{Conv}_{3 \times 3}(\mathcal{M}_2)$
\State Normalize $I$: 
$\hat{I} \gets \dfrac{I - \operatorname{min}(I)}{\operatorname{max}(I) - \operatorname{min}(I) + \varepsilon}$
\State Sample random $R \sim \mathcal{U}(\varepsilon, 1-\varepsilon)$
\State Threshold $\hat{I}$ with $R$ to get low-res mask $\mathcal{M}_{\text{final}}^{(\text{lowres})}$
\State Upsample to full resolution: 
$\mathcal{M}_{\text{final}} \gets \operatorname{Interp}_{H \times W}\!\bigl(\mathcal{M}_{\text{final}}^{(\text{lowres})}\bigr)$
\State \Return $\mathcal{M}_{\text{final}}$
\end{algorithmic}
\end{algorithm}

\section{Experiments}

To evaluate the effectiveness of GBGM, we conduct extensive experiments on three aspects:  
(1) image classification, to assess recognition improvements across different architectures and datasets;  
(2) image reconstruction, to examine the ability of GBGM to preserve semantic content and guide structure-aware recovery from masked inputs; and  
(3) empirical complexity analysis, to benchmark the runtime efficiency and scalability of GBGM against common augmentation strategies.

\subsection{Image Classification}

\subsubsection{CIFAR-10 with ResNet}
We evaluate GBGM on CIFAR-10 with six ResNet~\cite{he2016deep} backbones (20/32/44/56/110/1202).  
Given the small image size ($32\times32$), we adopt a single-stage mask: each image is split into $4\times4$ patches to form an $8\times8$ grid (Fig.~\ref{fig1}(a)).  
Two masking ratios, mask10 and mask20, remove the lowest 10\% and 20\% of patches, with a random\_mask10 baseline for comparison.  
Training follows standard CIFAR-10 settings: 200 epochs, batch size 128, SGD with a cosine-annealing schedule, and common augmentations (random crop, horizontal flip, normalization).

Table~\ref{tab:cifar10-results} reports Top-1 accuracy for all backbones.  
GBGM consistently outperforms the unmasked baseline and random masking: on ResNet-44, 10\% GBGM raises accuracy from 93.10\% to 94.68\%, while random masking drops it to 91.29\%.  
Across all six models, the average gain is about 0.8--1.0 percentage points at 10\% masking and slightly smaller at 20\%, showing robustness to both network depth and masking level. Unlike random masking, which may remove essential foreground information, GBGM steers attention to informative regions and encourages learning of global, complementary cues.  
The milder benefit at 20\% indicates that a moderate 10\% structural erasure provides the best balance between regularization and semantic preservation.  
By suppressing uninformative areas at the patch level, GBGM yields cleaner intermediate features and more compact representations, which in turn improve inter-class separability and stabilize training dynamics.  
These results confirm that GBGM offers a principled alternative to random masking, providing consistent and interpretable gains in accuracy.

Table~\ref{tab:cifar10-augment} compares GBGM with common augmentation methods on ResNet-44.  
While HAS reaches 94.35\% and others remain lower, GBGM achieves 94.68\%, matching or surpassing these strong baselines.  
Because GBGM operates on structural saliency rather than pixel-level distortions, it complements standard augmentations and can be combined with them for further gains.  
Overall, Table~\ref{tab:cifar10-augment} shows that GBGM delivers consistent accuracy improvements through structure-aware regularization, remains robust across network depths and masking ratios, and integrates naturally into standard training pipelines without extra hyper-parameters. This confirms that GBGM is both easy to integrate and consistently beneficial within standard CIFAR-10 training pipelines.  
Moreover, its structure-aware design suggests strong potential to generalize beyond CIFAR-10 to broader visual recognition tasks.

\begin{table}[h]
\centering
\caption{CIFAR-10 Top-1 accuracy (\%) of ResNet models with and without GBGM.}
\label{tab:cifar10-results}
\begin{tabular}{lcccc}
\toprule
\multirow{2}{*}{Model} & \multirow{2}{*}{Without Mask} & \multicolumn{2}{c}{With Mask} & \multirow{2}{*}{\makecell{Random\\Mask 10\%}} \\
 & & 10\% & 20\% & \\
\midrule
ResNet-20   & 92.86 & \textbf{92.99} & 92.80 & 91.62 \\
ResNet-32   & 92.63 & \textbf{93.94} & 93.91 & 92.46 \\
ResNet-44   & 93.10 & \textbf{94.68} & 93.94 & 91.29 \\
ResNet-56   & 93.62 & \textbf{94.44} & 93.68 & 92.31 \\
ResNet-110   & 93.57 & \textbf{94.67} & 94.54 & 93.59 \\
ResNet-1202  & 93.82 & \textbf{94.85} & 94.79 & 92.00 \\
\bottomrule
\end{tabular}
\end{table}

\begin{table}[h]
\centering
\caption{CIFAR-10 classification accuracy (\%) on ResNet-44 with different data augmentation methods.}
\label{tab:cifar10-augment}
\begin{tabular}{lc}
\toprule
Data Augmentation & Accuracy (\%) \\
\midrule
Baseline         & 93.10 \\
Random Erasing~\cite{zhong2020random}   & 92.65 \\
AutoAugment~\cite{cubuk2019autoaugment}      & 91.52 \\
GridMask~\cite{chen2020gridmask}         & 92.68 \\
HAS~\cite{singh2018hide}              & 94.35 \\
\midrule
GBGM (Ours) & \textbf{94.68} \\
\bottomrule
\end{tabular}
\end{table}

\begin{figure*}[t]
\centering
\includegraphics[width=0.92\textwidth]{fig/fig_cifar10_100.pdf}
\caption{(a) CIFAR-10 example and (b) CIFAR-100 example. The first row shows the original input image; the second row presents the partitioning result using the Granular Ball representation; the third row illustrates the saliency heatmap guided by granular balls; and the fourth row shows the final importance mask for image masking.}
\label{fig1}
\end{figure*}

\subsubsection{CIFAR-100 with EfficientNet-L2}

To further assess the generalization of GBGM, we conduct experiments on the CIFAR-100 dataset using EfficientNet-L2~\cite{2021Sharpness} as the backbone.  
Because CIFAR images are small ($32 \times 32$), all inputs are first resized to $224 \times 224$ to match the model input resolution.  
We train EfficientNet-L2 for 30 epochs on a single GPU using the SAM optimizer with cosine-annealing learning rate scheduling and a weight decay of $1\times10^{-5}$, while keeping all hyper-parameters identical across experiments.  
Standard data preprocessing includes resizing, random cropping, and horizontal flipping. Each baseline augmentation (AutoAugment, Random Erasing, GridMask, HAS) is tested independently for fair comparison. GBGM (Ours) is applied in a single stage similar to the CIFAR-10 experiments, as show in Fig.~\ref{fig1}(b).  
Each image is divided into non-overlapping $4 \times 4$ patches, and the 10\% least informative patches are masked according to our importance estimation before resizing to $224 \times 224$.  
All other training settings remain unchanged.

Table~\ref{tab:cifar100-effnet} summarizes the classification accuracy.  
GBGM achieves the best results, improving Top-1 accuracy from 94.79\% to 94.95\% and Top-5 accuracy from 99.30\% to 99.45\%.  
In contrast, conventional augmentations such as AutoAugment, Random Erasing, GridMask, and HAS lead to significantly lower performance, with Top-1 accuracy mostly below 93\% and Top-5 around 96–99\%.  
These results indicate that GBGM provides stable and non-trivial gains even on a strong SAM baseline.  
By explicitly guiding the network to focus on structurally salient regions while suppressing redundant background, GBGM supplies complementary regularization beyond pixel-level color perturbations, improving inter-class separability and overall robustness.  
Furthermore, its single-stage 10\% masking preserves semantic integrity and avoids the risk of randomly discarding informative patches, enabling nearly 99.5\% Top-5 accuracy without harming recognition of less dominant classes.  
This structural guidance also leads to more compact and discriminative feature representations, facilitating faster convergence during training.  
Notably, GBGM’s improvements remain consistent across different training seeds, underscoring its stability.

\begin{table}[H]
\centering
\caption{CIFAR-100 classification accuracy (\%) of EfficientNet-L2 (SAM) with different data augmentation methods.}
\label{tab:cifar100-effnet}
\begin{tabular}{lcc}
\toprule
Data Augmentation & Top-1 Acc. & Top-5 Acc. \\
\midrule
Baseline & 94.79 & 99.30 \\
Random Erasing~\cite{zhong2020random} & 87.00 & 97.01 \\
AutoAugment~\cite{cubuk2019autoaugment} & 92.95 & 99.17 \\
GridMask~\cite{chen2020gridmask} & 85.61 & 96.48 \\
HAS~\cite{singh2018hide} & 85.71 & 96.94 \\
\midrule
GBGM (Ours) & \textbf{94.95} & \textbf{99.45} \\
\bottomrule
\end{tabular}
\end{table}


\begin{figure*}[t]
\centering
\includegraphics[width=0.92\textwidth]{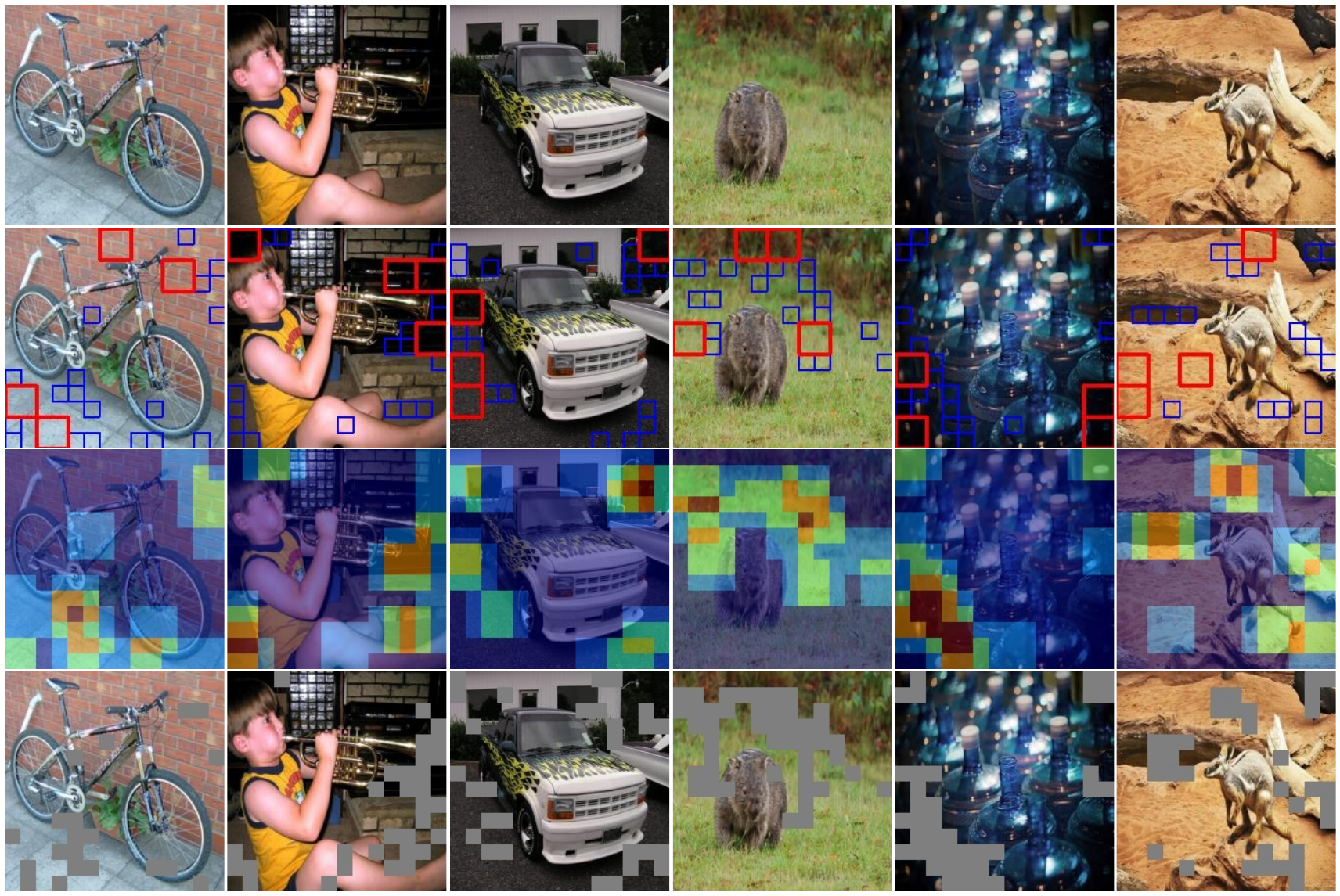}
\caption{Visualization of GBGM on ImageNet-1K.
From top to bottom: original images, Granular Ball partitioning, saliency heatmaps, and final importance masks.
The generated masks highlight semantically informative regions and suppress background, helping ViT focus on structurally important areas and improving classification accuracy.}
\label{fig:imagenet1k-mask}
\end{figure*}

\subsubsection{ImageNet-1K on CNN and Transformer Backbones}

We further evaluate GBGM on the ImageNet-1K dataset with two representative architectures: a convolutional network (EfficientNet-B0) and a hierarchical vision transformer (Swin Transformer-Tiny, Swin-T)~\cite{liu2021swin}, both implemented via the \texttt{timm} library. For EfficientNet-B0, we train for 30 epochs on a single NVIDIA GPU with batch size 64, using AdamW (learning rate $1\times10^{-4}$, weight decay $1\times10^{-5}$) and a cosine-annealing learning-rate schedule. For Swin-T, we fine-tune the pretrained model on a single GPU with AdamW and a warmup+cosine-annealing schedule. In all cases, the input resolution is fixed to $224\times224$, and the data pipeline follows standard ImageNet preprocessing: resize to $224\times224$, random horizontal flip, mild color jitter/small rotation, and standard ImageNet normalization. For each backbone, we vary only the data augmentation strategy (AutoAugment, Random Erasing, GridMask, HAS), keeping all other hyper-parameters identical.

Across both architectures, GBGM differs solely in the masking strategy while keeping all optimization settings unchanged. For a $224\times224$ input, stage~1 partitions the image into $32\times32$ blocks (a $7\times7$ grid) and removes the lowest-scoring 10\% based on a block-level purity measure. Stage~2 refines the selection within the remaining regions using $16\times16$ blocks (a $14\times14$ grid), again removing the lowest 10\%. This yields a binary importance mask at $16\times16$ granularity, which is resized back to the input resolution and applied multiplicatively to the image. The mask is only used during training; at inference time, both EfficientNet-B0 and Swin-T operate on the original images without masking.

Table~\ref{tab:imagenet1k-results} reports the validation accuracy across augmentation strategies and backbones. On EfficientNet-B0, GBGM improves the Top-1 / Top-5 accuracy from 74.31\% / 92.27\% (Baseline) to 74.68\% / 92.60\%, achieving the best performance among all compared augmentations. On Swin Transformer-Tiny, classical augmentations again provide only marginal gains over the 80.31\% / 95.11\% Baseline (e.g., GridMask reaches 80.36\% / 95.18\%), whereas GBGM yields the highest accuracy of \textbf{80.78\%} Top-1 and \textbf{95.27\%} Top-5. These consistent improvements on both a CNN and a transformer backbone indicate that structure-guided masking offers a generic and effective regularization mechanism: by hierarchically suppressing redundant, homogeneous background regions while preserving object-centric and boundary-sensitive details, GBGM provides saliency-driven regularization that avoids randomly discarding informative content and enhances generalization on large-scale classification tasks.

\begin{table}[t]
    \centering
    \caption{ImageNet-1K validation accuracy (\%) on EfficientNet-B0 (Eff-B0) and Swin Transformer-Tiny (Swin-T) with different data augmentation strategies.}
    \label{tab:imagenet1k-results}
    \begin{tabular}{lcccc}
        \toprule
        \multirow{2}{*}{Method} & \multicolumn{2}{c}{Eff-B0} & \multicolumn{2}{c}{Swin-T} \\
        \cmidrule(lr){2-3} \cmidrule(lr){4-5}
        & Top-1 & Top-5 & Top-1 & Top-5 \\
        \midrule
        Baseline                         & 74.31 & 92.27 & 80.31 & 95.11 \\
        Random Erasing~\cite{zhong2020random}  & 74.58 & 92.41 & 80.33 & 95.12 \\
        AutoAugment~\cite{cubuk2019autoaugment} & 73.34 & 91.64 & 80.33 & 95.14 \\
        GridMask~\cite{chen2020gridmask}       & 73.34 & 91.64 & 80.36 & 95.18 \\
        HAS~\cite{singh2018hide}               & 74.27 & 92.17 & 80.35 & 95.14 \\
        \midrule
        \textbf{GBGM (ours)}                   & \textbf{74.68} & \textbf{92.60} & \textbf{80.78} & \textbf{95.27} \\
        \bottomrule
    \end{tabular}
\end{table}

\subsection{Image Reconstruction}

To further evaluate GBGM for large-scale pretraining, we integrate it into the Masked Autoencoder (MAE)~\cite{he2022masked} with a ViT-Base backbone and train on ImageNet-100 (100 random classes from ImageNet-1K, resized to $224\times224$). Our hierarchical masking uses patch sizes of $32 \times 32$ and $16 \times 16$.  

\begin{figure}[htbp]
    \centering
    \includegraphics[width=0.85\columnwidth]{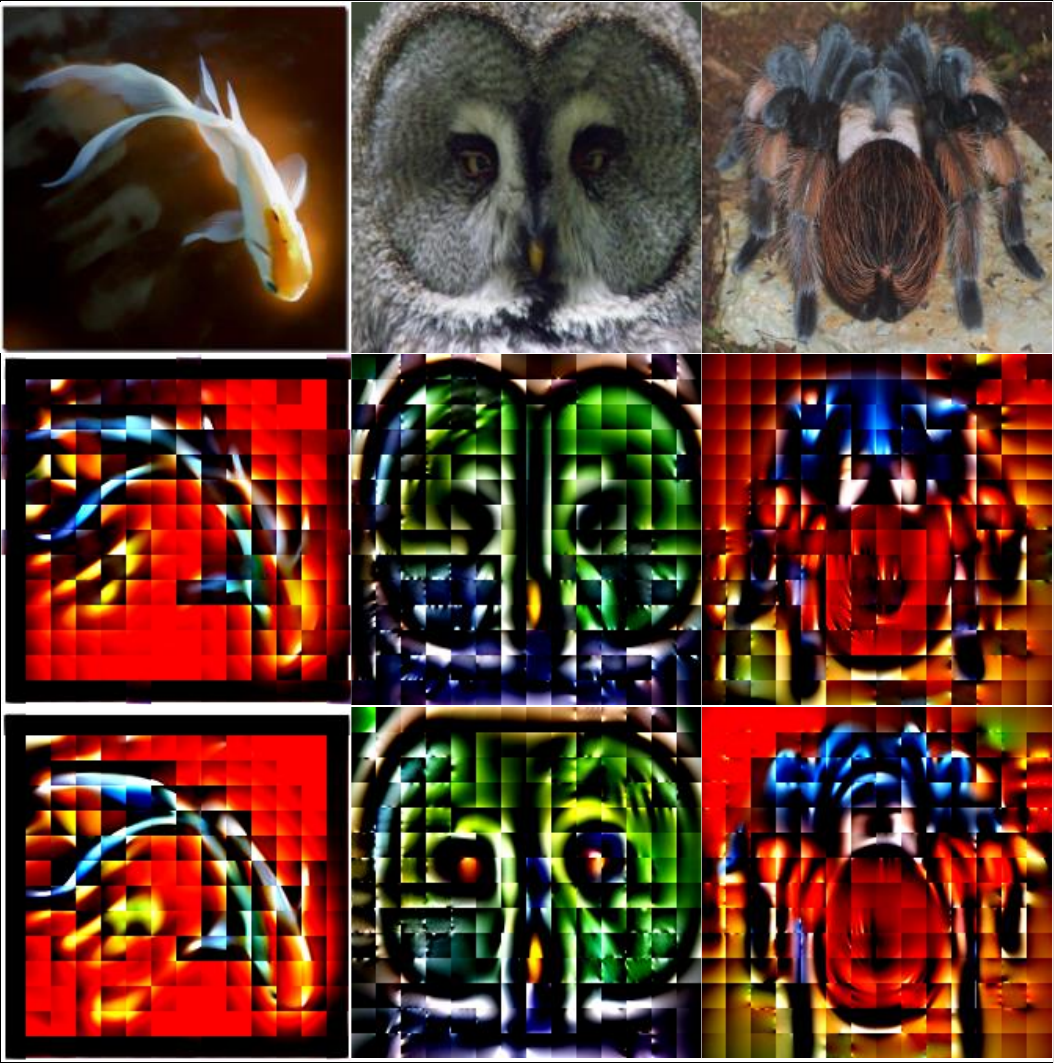}
    \caption{Reconstruction comparison on ImageNet-100. 
    Top: original images. Middle: reconstructions by MAE. 
    Bottom: reconstructions by GBGM-enhanced MAE.}
    \vspace{-5pt} 
    \label{fig:imagenet100-recon}
\end{figure}

Table~\ref{tab:mae-recon} reports PSNR, SSIM, and LPIPS. Compared with vanilla MAE, GBGM achieves slightly higher PSNR/SSIM and lower LPIPS, indicating improved perceptual fidelity. Visual results in Fig.~\ref{fig:imagenet100-recon} further show that high-frequency details (e.g., fish heads, owl eyes, spider contours) are more sharply reconstructed, whereas MAE often exhibits edge blurring. By directing attention to structurally salient regions, GBGM enhances both local texture recovery and global shape consistency, yielding reconstructions that are visually sharper and semantically more accurate.

\begin{table}[h]
\centering
\caption{Image reconstruction performance on ImageNet-100.}
\label{tab:mae-recon}
\begin{tabular}{lccc}
\toprule
Method & PSNR ↑ & SSIM ↑ & LPIPS ↓ \\
\midrule
MAE~\cite{he2022masked} & 8.22 dB & 0.231 & 0.640 \\
GBGM (Ours) & \textbf{8.26 dB} & \textbf{0.252} & \textbf{0.623} \\
\bottomrule
\end{tabular}
\end{table}

\subsection{Image Tampering Detection}
\label{subsec:tamper_detection}

Image tampering detection aims to determine whether an image has been manipulated and to localize the tampered regions. This task is particularly challenging when the test distribution differs from the training data due to changes in manipulation sources, post-processing operations, or imaging conditions. To validate the effectiveness of GBGM beyond classification, we follow the ADCD-based tampering detection protocol in~\cite{wong2025adcd} and build our detector upon the same baseline model. Specifically, we adopt ADCD as the baseline and plug in GBGM to form ADCDNet\_GBGM, and we further compare with several representative augmentation-based variants, including Random Erasing~\cite{zhong2020random}, AutoAugment~\cite{cubuk2019autoaugment}, GridMask~\cite{chen2020gridmask}, and HAS~\cite{singh2018hide}. Table~\ref{tab:tamper_detection} reports the quantitative results on three validation sets, i.e., TestSet, FCD, and SCD. Overall, ADCDNet\_GBGM achieves the highest average F1 of 91.36\%, improving upon the baseline (90.30\%) by 1.06 points, and also outperforming Random Erasing (90.71\%), AutoAugment (90.71\%), GridMask (90.58\%), and HAS (90.41\%). On the more challenging SCD set, GBGM increases F1 from 86.27\% to 87.77\% (+1.50), while also improving F1 on TestSet (from 91.97\% to 92.65\%, +0.68) and FCD (from 92.65\% to 93.66\%, +1.01). Notably, HAS attains the highest recall across all three sets, whereas GBGM yields the best precision and F1, indicating that GBGM improves overall detection quality under both in-domain and cross-domain evaluation settings.

\begin{table}[t]
\centering
\caption{Performance comparison on different validation sets (\%).}
\label{tab:tamper_detection}
\renewcommand{\arraystretch}{1.1}
\setlength{\tabcolsep}{4pt}
\resizebox{\linewidth}{!}{%
\begin{tabular}{l l c c c c}
\hline
Method & Val Set & Pre & Recall & F1 & Avg. F1 \\
\hline
\multirow{3}{*}{Baseline~\cite{wong2025adcd}} & TestSet & 90.16 & 93.86 & 91.97 & \multirow{3}{*}{90.30} \\
                                             & FCD     & 89.47 & 96.07 & 92.65 & \\
                                             & SCD     & 79.93 & 93.71 & 86.27 & \\
\hline
\multirow{3}{*}{Random Erasing~\cite{zhong2020random}} & TestSet & 89.76 & 95.00 & 92.31 & \multirow{3}{*}{90.71} \\
                                                      & FCD     & 91.39 & 95.13 & 93.23 & \\
                                                      & SCD     & 79.09 & 95.67 & 86.59 & \\
\hline
\multirow{3}{*}{AutoAugment~\cite{cubuk2019autoaugment}} & TestSet & 89.76 & 95.00 & 92.31 & \multirow{3}{*}{90.71} \\
                                                        & FCD     & 91.39 & 95.13 & 93.23 & \\
                                                        & SCD     & 79.09 & 95.67 & 86.59 & \\
\hline
\multirow{3}{*}{GridMask~\cite{chen2020gridmask}} & TestSet & 90.13 & 94.29 & 92.17 & \multirow{3}{*}{90.58} \\
                                                 & FCD     & 90.48 & 95.55 & 92.95 & \\
                                                 & SCD     & 79.83 & 94.66 & 86.62 & \\
\hline
\multirow{3}{*}{HAS~\cite{singh2018hide}} & TestSet & 86.23 & \textbf{95.97} & 90.84 & \multirow{3}{*}{90.41} \\
                                         & FCD     & 88.80 & \textbf{97.11} & 92.77 & \\
                                         & SCD     & 80.40 & \textbf{96.31} & 87.64 & \\
\hline
\multirow{3}{*}{GBGM (Ours)} & TestSet & \textbf{91.15} & 94.19 & \textbf{92.65} & \multirow{3}{*}{\textbf{91.36}} \\
                            & FCD     & \textbf{93.54} & 93.78 & \textbf{93.66} & \\
                            & SCD     & \textbf{81.52} & 95.06 & \textbf{87.77} & \\
\hline
\end{tabular}%
}
\end{table}

\subsection{Empirical Complexity Analysis}

While the previous subsection provided a theoretical analysis showing that GBGM has linear complexity $O(HW)$ with respect to image resolution, we now turn to empirical measurements to assess its practical runtime behavior. We benchmarked five augmentation strategies—Random Erasing, AutoAugment, GridMask, HAS, and GBGM (Ours)—on a CUDA device at $224\times224$, varying batch size from 1 to 32 and repeating each setting 15 times. Average time, throughput, and a scaling factor $\beta$ were measured via log-log regression, where $\beta \approx 1$ denotes linear scaling and $\beta < 1$ indicates sub-linear behavior.

Results in Table~\ref{tab:time-complexity} show Random Erasing is fastest and HAS slowest, with AutoAugment/GridMask moderate. GBGM averages $3.78$ms and 2256.3 samples/sec. Table~\ref{tab:scalability} further shows $\beta=0.601$, meaning that per-sample cost decreases as batch size grows. This observation is consistent with our theoretical analysis: although GBGM has linear complexity $O(HW)$ with respect to image resolution, its mask generation pipeline is highly parallelizable, allowing GPUs to amortize computation across larger batches. As a result, GBGM achieves favorable sub-linear scalability in practice, making it particularly suitable for large-scale parallel training.

\begin{table}[ht]
\centering
\caption{Runtime performance of augmentation methods (ms and samples/sec).}
\label{tab:time-complexity}
\begin{tabular}{lcccc}
\toprule
Method & Min & Max & Avg & Throughput \\
\midrule
Random Erasing~\cite{zhong2020random} & 0.14 & 3.70  & 1.21  & 8348.3 \\
AutoAugment~\cite{cubuk2019autoaugment} & 0.82 & 20.51 & 6.65  & 1547.6 \\
GridMask~\cite{chen2020gridmask} & 0.47 & 13.59 & 4.52  & 2322.3 \\
HAS~\cite{singh2018hide} & 1.30 & 39.12 & 12.87 & 810.3  \\
\midrule
GBGM (Ours) & 1.30 & 9.98  & 3.78  & 2256.3 \\
\bottomrule
\end{tabular}
\end{table}

\begin{table}[ht]
\centering
\caption{Scalability factors of different augmentation methods.}
\label{tab:scalability}
\begin{tabular}{lcc}
\toprule
Method & Scaling Factor & Trend \\
\midrule
Random Erasing~\cite{zhong2020random} & 0.950 ($R^2=0.997$) & Linear  \\
AutoAugment~\cite{cubuk2019autoaugment} & 0.956 ($R^2=0.993$) & Linear  \\
GridMask~\cite{chen2020gridmask} & 0.994 ($R^2=0.998$) & Linear \\
HAS~\cite{singh2018hide} & 0.991 ($R^2=1.000$) & Linear  \\
\midrule
GBGM (Ours) & 0.601 ($R^2=0.907$) & Sub-linear \\
\bottomrule
\end{tabular}
\end{table}

\section{Conclusion}

We presented Granular Ball Guided Masking (GBGM), a structure-aware augmentation strategy that integrates Granular Ball Computing with hierarchical masking. By organizing image regions into a coarse-to-fine structure, GBGM preserves semantically informative and structurally important content while suppressing redundant areas, producing discriminative masks that can be seamlessly applied to both CNNs and Vision Transformers. Extensive experiments on CIFAR-10/100, ImageNet-100, and ImageNet-1K demonstrate consistent gains in classification accuracy and favorable sub-linear scalability. Beyond recognition, GBGM also improves performance on image tampering detection under the ADCD-based setting, suggesting that structure-aware masking can benefit forensic robustness under distribution shifts. Future work will explore extending GBGM to self-supervised pretraining and dense prediction tasks, where structure-aware masking may further support spatial and temporal consistency.

\bibliography{ref}
\bibliographystyle{IEEEtran}

\end{document}